\documentclass[conference]{IEEEtran}
\IEEEoverridecommandlockouts
\usepackage{cite}
\usepackage{amsmath,amssymb,amsfonts}
\usepackage{algorithmic}
\usepackage{graphicx}
\usepackage{textcomp}
\usepackage{xcolor}

\usepackage{booktabs}

\def\BibTeX{{\rm B\kern-.05em{\sc i\kern-.025em b}\kern-.08em
    T\kern-.1667em\lower.7ex\hbox{E}\kern-.125emX}}
\begin{document}

\title{SOV-CAD: Stepwise Orthographic Views Guided CAD Modeling Sequence Reconstruction}

% \author{Anonymous ICME submission}
% \author {
%     % Authors
%     Zhaopeng Feng\textsuperscript{\rm 1},
%     Chen Zhi\textsuperscript{\rm 1},
%     Xuhong Zhang\textsuperscript{\rm 1},
%     Zhengwen Feng\textsuperscript{\rm 1},
%     Xinkui Zhao\textsuperscript{\rm 1}
    
% }

% \IEEEauthorblockN {
%     % Affiliations
% {\rm 1}School of Software Technology, Zhejiang University 
%     % \textsuperscript{\rm 2}Affiliation 2\\
%     % \{zpfeng, zjuzhichen, zhangxuhong, fengzhengwen, zhaoxinkui\}@zju.edu.cn
% }

\author{\IEEEauthorblockN{Zhaopeng Feng\textsuperscript{\rm 1},
    Chen Zhi\textsuperscript{\rm 1, 2, $\dagger$},
    Xuhong Zhang\textsuperscript{\rm 1, 2},
    Zhengwen Feng\textsuperscript{\rm 1, 3},
    Xinkui Zhao\textsuperscript{\rm 1, 2}}
\IEEEauthorblockA{
\textit{\textsuperscript{\rm 1}School of Software Technology, Zhejiang University} \\
\textit{\textsuperscript{\rm 2}Zhejiang Key Laboratory of Digital-Intelligence Service Technology, Zhejiang University} \\
\textit{\textsuperscript{\rm 3}ZWSOFT Co., Ltd.}\\
\thanks{\textsuperscript{$\dagger$}\,Corresponding author: Chen Zhi (zjuzhichen@zju.edu.cn)}
% \   {name of organization (of Aff.)}\\
% City, Country \\
\{zpfeng, zjuzhichen, zhangxuhong, fengzhengwen, zhaoxinkui\}@zju.edu.cn
}
% \thanks{\textsuperscript{$\dagger$}\,Corresponding author: Chen Zhi (zjuzhichen@zju.edu.cn)}
}

\maketitle

\begin{abstract}
Reconstructing Computer-Aided Design (CAD) modeling sequences from images is crucial for preserving design intent and supporting parametric editing. However, existing methods typically generate full CAD sequences holistically, overlooking the iterative, feedback-driven nature of human design workflows. We address this limitation by introducing the rich \emph{stepwise visual supervision}: at each modeling step, the system observes the target’s orthographic projections, the projections of the incrementally constructed model, and the active sketch, enabling informed action selection. To effectively leverage this on-the-fly feedback, we propose SOV-CAD, a framework that formulates CAD reconstruction as a sequential decision-making task and employs offline reinforcement learning with a Decision Transformer architecture. This design incorporates continuous visual feedback guided by geometric alignment rewards, resulting in a more accurate and human-like modeling process. Extensive experiments show that SOV-CAD surpasses state-of-the-art methods in CAD sequence reconstruction while exhibiting strong data efficiency. Code of SOV-CAD is available at: https://github.com/LukePhong/SOV-CAD

\end{abstract}


\section{Introduction}
\label{sec:intro}

Reconstructing the procedural modeling sequences of Computer-Aided Design (CAD) models from images is an increasingly important goal in intelligent design automation. Unlike static 3D reconstruction, recovering the underlying parametric operations preserves design intent, supports editable and manufacturable geometry, and enables downstream reuse \cite{Willis2021Fusion360Gallery}. However, translating visual inputs into structured CAD programs remains challenging due to intrinsic 2D–3D ambiguities, the large combinatorial search space of modeling operations, and variations in both imaging conditions and design styles \cite{You2024Img2CAD}.

\begin{figure}[htbp]
    \centerline{\includegraphics[width=0.9\columnwidth]{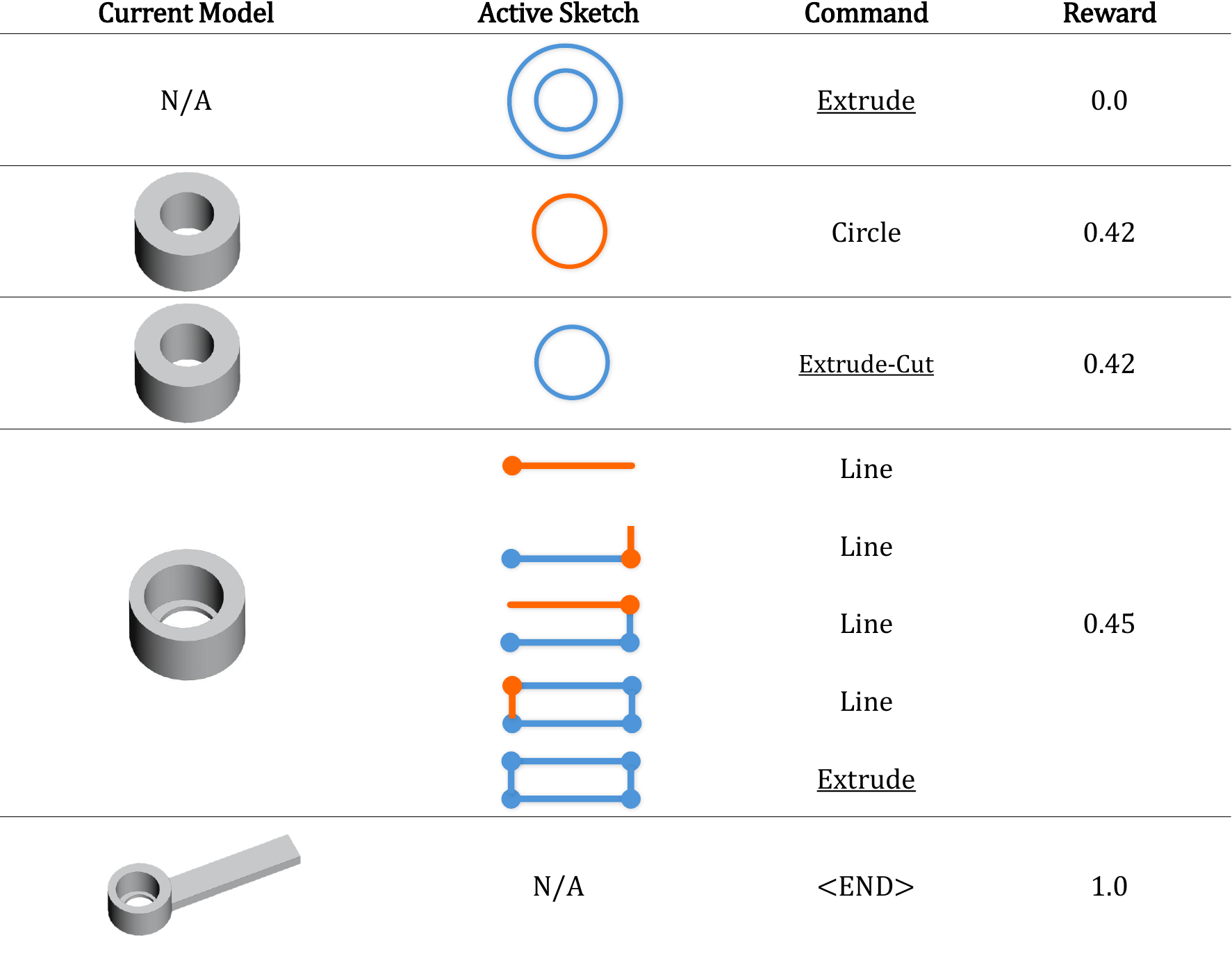}}
    \caption{An illustration of the step-by-step CAD model construction process, which applies to both human CAD experts and our method. Modeling commands include sketch operations and various feature operations (underlined). The modeler draws the active sketch to outline the feature, and then uses feature operations to modify the 3D solid. Reward reflects the similarity between the current model and the target, which can be calculated in different ways to mimic the human CAD experts estimating their completeness.}
    \label{fig:cad_step}
\end{figure}

Recent deep-learning-based methods have achieved notable success in reconstructing 3D shapes from images \cite{long2023wonder3d}, yet they primarily generate unstructured meshes that lack the precision and editability required in CAD. Conversely, methods that aim to produce structured CAD representations typically generate the entire modeling sequence or Boundary Representation (B-Rep) holistically \cite{liu2025hola, Li2025MambaCAD, Chen2024CADCrafter}. Such holistic approaches treat the modeling sequence as indivisible, rely on high-dimensional latent encodings and overlook the inherently stepwise nature of CAD modeling (Fig.~\ref{fig:cad_step}), limiting opportunities for intermediate feedback and leading to inefficient learning and optimization.

% Existing approaches to generate structured CAD models still face significant limitations. As shown in Fig.~\ref{fig:cad_step}, the CAD modeling process is essentially stepwise. 

% Furthermore, existing approaches to generate structured CAD models typically produce the entire sequence or Boundary Representation (B-Rep) holistically \cite{liu2025hola, wu2025cmt, Li2025MambaCAD, Chen2024CADCrafter}. By overlooking the inherently stepwise nature of CAD modeling (Fig.~\ref{fig:cad_step}), these holistic approaches \textbf{miss the opportunity to obtain stepwise feedback during generation}, resulting in enormous solution spaces that complicate both learning and inference, longer processing pipelines, and fewer opportunities for optimization and error correction. 

% For example, BrepGen \cite{Xu2024BrepGen} represents B-rep models through structured latent geometry organized in hierarchical trees, while other methods, such as CADCrafter \cite{Chen2024CADCrafter} and MambaCAD \cite{Li2025MambaCAD}, generate complete operation sequences directly from latent representations.

A key observation is that CAD modeling is inherently \emph{multimodal}: it involves parametric operation sequences while relying heavily on visual evaluations of the developing geometry \cite{wang2025Text-to-CAD}. Human designers continuously refer to intermediate visual feedback to assess correctness and guide subsequent decisions. However, existing approaches use visual information only for post-hoc validation or refinement, rather than integrating it into the modeling loop itself \cite{wang2025Text-to-CAD, alrashedy2025CADCodeVerify, Li2025SeekCAD}. This gap prevents current systems from emulating expert design behavior.

To address these limitations, we propose \textbf{SOV-CAD}, a framework that reconstructs CAD modeling sequences by observing the target geometry and acting step-by-step. At each step, the system receives on-the-fly visual feedback through orthographic three-view projections of the incrementally constructed model and the evolving sketch canvas. Together with a stepwise geometric consistency reward, these multimodal supervisions provide precise intermediate guidance analogous to the visual reasoning of human designers.

% In this way, our method differs from previous work that uses visual feedback only for post-generation refinement \cite{Li2025SeekCAD}.  

Motivated by insights from Fusion 360 \cite{Willis2021Fusion360Gallery}, we cast CAD reconstruction as a sequential decision-making problem. Instead of generating long sequences holistically, we decompose the process into a series of manageable decisions. We adopt reinforcement learning (RL) to leverage the state and reward of the CAD modeling environment. Because interacting with CAD software to obtain online feedback is impractical, we adopt an offline RL setting and employ a Decision Transformer (DT) \cite{Chen2021DecisionTransformer}. Prior studies demonstrate the scalability and effectiveness of Transformer-based models for CAD sequence generation \cite{Wu2021DeepCAD, Xu2022SkexGen, Zhou2023CADParser}. Building on this foundation, we introduce—\textit{to the best of our knowledge—the first method that integrates Decision Transformers with stepwise visual feedback for CAD sequence recovery}. The model learns to select operations that maximize geometric consistency with the target design.

% Our carefully designed model achieves superior results with better data efficiency compared to the state-of-the-art approaches, with excellent generalization capabilities on both mesh-represented 3D models and real-world images.
% The experiments on both mesh-represented 3D models and real-world images demonstrate that this method exhibits excellent generalization capabilities.

Our contributions are summarized as follows:
\begin{itemize}
\item \textbf{Novel stepwise multimodal supervision.} We introduce orthographic three-view projections of intermediate CAD states, combined with sketch canvases, as dynamic visual supervisory signals that exploit the step-by-step nature of CAD modeling.
\item \textbf{Practical Decision Transformer framework for CAD reconstruction.} We formulate CAD modeling as an offline RL task and design a DT-based framework that conditions on past states, actions, and rewards to reduce decision complexity and improve robustness.
\item \textbf{State-of-the-art reconstruction performance.} Our method achieves superior accuracy and strong data efficiency in recovering CAD modeling sequences, with excellent generalization ability.
\end{itemize}

\section{Related Work}

\subsection{CAD Sequence Modeling}
% Modeling CAD generation as a sequential process, analogous to natural language, has become a major research focus. 

DeepCAD \cite{Wu2021DeepCAD} first showed that Transformer-based autoencoders can learn latent spaces for modeling sequences. Subsequent work enhanced control via autoregressive models with disentangled or hierarchical representations \cite{Xu2022SkexGen, Xu2023HNCCAD}, or generated sequences holistically from latent spaces recovered through diffusion \cite{Chen2024CADCrafter}. This paradigm has been extended to reverse engineering \cite{Zhou2023CADParser, Li2025MambaCAD} and, more recently, to fine-tuned LLMs for CAD sequence modeling \cite{ Wang2025From2DCAD, Zhang2025FlexCAD}.
However, most methods generate sequences holistically and lack the iterative feedback that is central to human design. 

% Our work addresses this by introducing an offline RL framework where an agent reconstructs CAD models step-by-step, guided by rich visual feedback.

\subsection{Reconstruction from Orthographic Views and Sketches}

Inferring 3D geometry from 2D projections is a long-standing challenge. Recent advances use deep learning on new datasets, such as PlankAssembly’s synthesis from orthographic views \cite{Hu2023PlankAssembly} or VLM-based translation from technical drawings \cite{Wang2025From2DCAD}.
Although effective in narrow domains, their inputs are difficult to source and often cannot generalize further.

% Early methods relied on deterministic geometric reasoning over vectorized drawings \cite{Idesawa1973}. 

\subsection{Reconstruct from Images}
Recovering CAD from images is hindered by the inherent 2D-to-3D ambiguity. Diffusion-based methods generate full CAD programs directly \cite{Chen2024CADCrafter, Li2025CADDreamer}, while LLMs/VLMs have been adapted to predict high-level structures for parameter regression like in Img2CAD \cite{You2024Img2CAD}, \cite{alrashedy2025CADCodeVerify} and \cite{mallis2025cadassistant} or to output programs in an end-to-end fashion \cite{Xu2024CADMLLM, Wang2025CADGPT}. 
However, visual feedback is typically post-hoc, used only for refinement \cite{Li2025SeekCAD}, rather than integrated step-by-step during modeling.

% Other approaches focus on specific primitive types or RL-based refinement \cite{Hong2024MV2Cyl, kolodiazhnyi2025cadrille}.

% \subsection{Reinforcement Learning for CAD Reconstruction}
% Reinforcement Learning has been explored for CAD reconstruction in various forms. RLCAD \cite{Yin2025RLCAD} trains agents in simulated environments but relies on B-Rep graph states, while \cite{Zhang2025RLReconstruction} recover B-Rep models from three-view sketches but require rare vectorized inputs. Other methods fine-tune LLMs/VLMs with RL algorithms such as DPO and GRPO \cite{Guan2025CADCoder, wang2025Text-to-CAD, Niu2025CReFTCAD}.
% In contrast, our method casts CAD reconstruction as a sequential decision-making problem with offline RL, leveraging intermediate geometric states for stepwise visual guidance.

\begin{figure*}[t]
    \centering
    \includegraphics[width=0.9\textwidth]{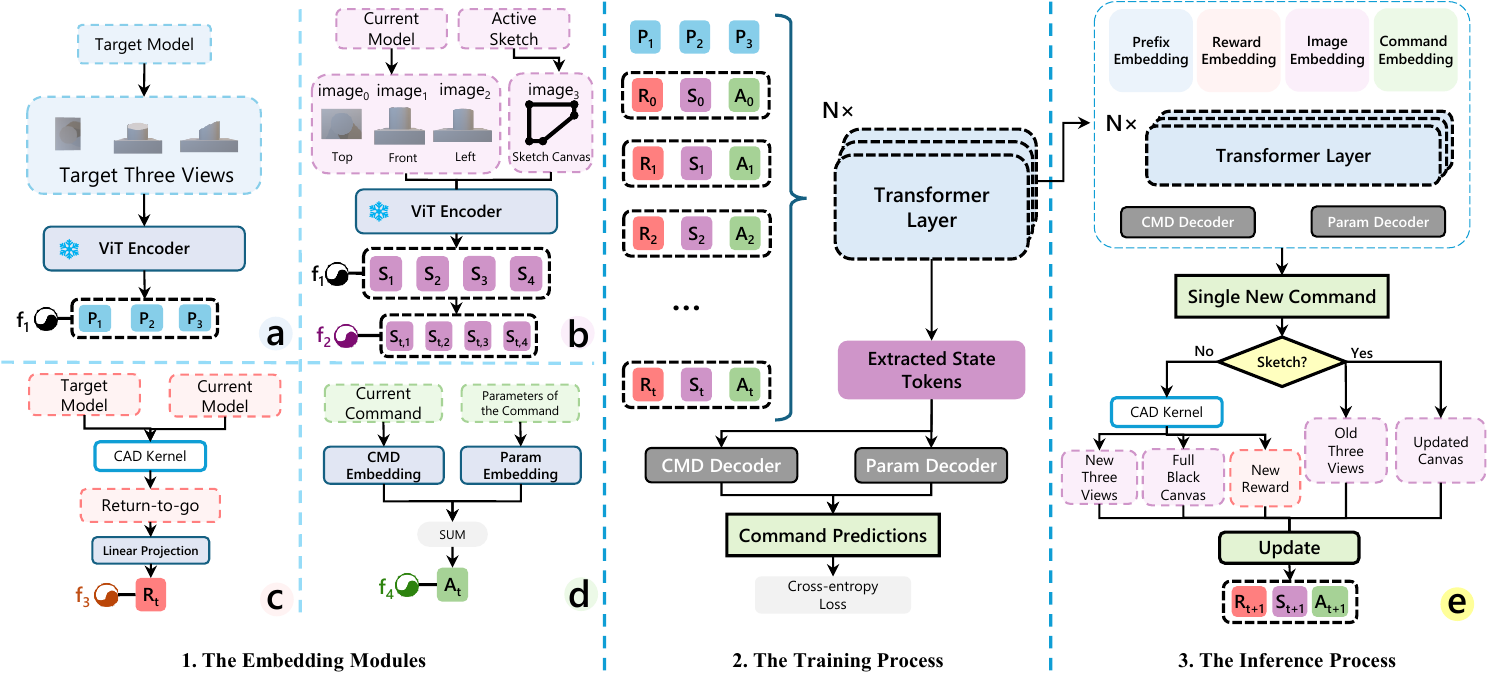} % Replace with actual figure
    % \fbox{\rule[-.5cm]{0cm}{8cm} \rule[-.5cm]{16cm}{0cm}}
    \caption{An overview of our model architecture. (a) The target model's three orthographic views are encoded into prefix tokens. (b) At each step $t$, the current model's three views and the active sketch are encoded into state tokens. (c) Convert the IoU between the current and the target model to the return-to-go token. (d) The CAD command is encoded into an action token. In the training process, the prefix tokens and the trajectory triplets are processed by a Transformer, which uses MLP decoders to predict actions. (e) In the inference process, the environment is updated based on the generated action, yielding new state and reward for step $t+1$. (f) Four types of positional embeddings providing context: for image tokens (f1), for state, return-to-go and action tokens (f2, f3, f4).}
    \label{fig:model_architecture}
\end{figure*}

\section{Methodology}

%CAD experts often begin with a mental concept or reference schematics, such as orthographic drawings, and iteratively refine their designs by experimenting with modeling commands and parameters. This process is guided by continuous visual inspection of the intermediate results. Our approach emulates this iterative workflow by incorporating stepwise visual feedback throughout the modeling process and guiding its own actions by automatically extracting geometric and semantic information from the images. Specifically, our framework integrates multiple information sources: orthographic views of the target model ($M_{target}$), orthographic projections of the incrementally constructed model ($M_{t}$), rasterized sketch canvas, running reward and evolving CAD operation sequence. The overall architecture is shown in Fig.~\ref{fig:model_architecture}.

In this section, we first describe the problem formulation, then introduce our offline RL method, and the architecture of our model, including how each modality is embedded and organized, as well as the training and inference processes.

CAD experts often begin with a mental concept or reference schematics, such as orthographic drawings. They iteratively refine their designs and experiment with modeling commands and parameters, guided by continuous visual inspections of the intermediate results. Our approach emulates this iterative workflow, and we formalize the problem as follows.
% Decision Transformer is an offline RL framework for visual states, and our method can be formalized as follows.
% We formalize the decision-making process as a conditional sequence modeling problem. 
Given a dataset $\mathcal{D}$ of CAD construction trajectories, our objective is to optimize the parameters $\theta$ of a policy network to maximize the likelihood of expert actions:
\begin{equation}
\theta^* = \operatorname*{argmax}_{\theta} \mathbb{E}_{\tau \sim \mathcal{D}} \left[ \sum_{t=0}^{N_c-1} \log P_\theta(a_t \mid T_p, H_{< t}) \right],
\end{equation}
where $\tau$ represents a trajectory sequence, $N_c$ is the sequence length, $T_p$ denotes the target representation, and $H_{t}$ represents the stepwise histories.
%Specifically, our framework integrates multiple information sources: orthographic views of the target model ($M_{target}$), orthographic projections of the incrementally constructed model ($M_{t}$), rasterized sketch canvas, running reward and evolving CAD operation sequence. 

% The architecture of our model is shown in Fig.~\ref{fig:model_architecture}.

\subsection{Offline Reinforcement Learning}

The CAD construction process is a natural fit for a Markov Decision Process, where an agent iteratively applies modeling operations to a geometry \cite{Willis2021Fusion360Gallery}. We further formalize this task as an offline reinforcement learning problem, avoiding the need for direct interaction with the CAD software in training.

In our RL formulation, the \textbf{state ($s_t$)} at each step $t$ includes the orthographic views of the current model ($V_{c,t}$) and the current sketch canvas ($V_{sk,t}$), providing the stepwise visual feedback. An \textbf{action ($a_t$)} is a discretized CAD operation with associated parameters. The \textbf{reward ($r_t$)} is defined as the improvement in the Intersection over Union (IoU) after applying the action $a_t$, $r_t = \mathrm{IoU}(M_{t}, M_{target}) - \mathrm{IoU}(M_{t-1}, M_{target})$, which provides progress in geometric fidelity between the current model ($M_{t}$) and the target model ($M_{target}$).
The objective is to learn a policy that maximizes the reward, thereby generating an accurate reconstruction sequence.

% The \textbf{reward ($r_t$)} provides the progress in geometric fidelity between the current model ($M_{t}$) and the target model ($M_{target}$), defined as the improvement in the Intersection over Union (IoU) after applying the action $a_t$:
% \begin{equation}
% r_t = \mathrm{IoU}(M_{t}, M_{target}) - \mathrm{IoU}(M_{t-1}, M_{target}).    
% \end{equation}
% contains all the visual information for the next decision, 
% providing progress in geometric fidelity between the current model ($M_{t}$) and the target model ($M_{target}$) and guides the reconstruction. 
 % $\pi(a_{t+1} | s_t)$ 

\subsection{Model Architecture and Input Representation}

% The architecture of SOV-CAD is shown in Fig.~\ref{fig:model_architecture}. 
To effectively learn such a policy from offline data, we draw inspiration from the Decision Transformer architecture. It models the offline RL problem as a conditional sequence modeling task, predicting future actions based on a sequence of history states, actions, and returns-to-go (sum of future rewards), with a GPT-like Transformer \cite{Chen2021DecisionTransformer}. As shown in Fig.~\ref{fig:model_architecture}, the input to the Transformer is a sequence of trajectory triplets $(\hat{R_t}, S_t, A_t)$. Each triplet consists of a return-to-go token $\hat{R_t}$, a state token group $S_t$, and an action token $A_t$. The DT framework accepts single-view input, which is insufficient. However, requiring too many views would obscure other supervisions and be inconvenient for users, so we made a trade-off and adopted the orthogonal three-view (front, top, left) of $M_{target}$ and $M_{t}$ commonly used in many studies \cite{Hu2023PlankAssembly, Zhang2025RLReconstruction}. 
Orthographic views also fix the positional relationship between viewpoints and simplify the problem.

% Orthographic views also fix the positional relationship between viewpoints, simplifying the problem.    The original DT framework only supports single-view input, which is insufficient.

\paragraph{Image Encoding}
% The model’s geometric understanding derives from visual inputs, which are processed using a unified tokenization pipeline. 

% This yields three tokens, each added to a unique learned positional embedding $P_{image,i}$ to preserve its identity (Fig.~\ref{fig:model_architecture}f1). These tokens are prepended to the input sequence to guide generation. 

% While three orthographic views may not capture all the geometric information, this choice allows us to explore the feasibility of a purely visual reconstruction pipeline, without explicit structural information as input.

Visual inputs are processed by a unified tokenization pipeline, allowing the model to efficiently relate the target geometry to its incremental progress within a consistent representation. Each image is encoded by a pretrained Vision Transformer (ViT) \cite{Dosovitskiy2020ViT}, and the resulting ``[CLS]'' embedding (extracted before the classification head) is linearly projected to the model dimension $d_{m}$ and layer-normalized. This produces two categories of tokens: prefix tokens and state tokens.

\textbf{Prefix tokens ($T_{p}$)} provide a fixed reconstruction target (Fig.~\ref{fig:model_architecture}a). They are generated from the three orthographic views of the complete target model, $M_{target}$. This yields three tokens, and we prepend them to the input sequence throughout the process to guide generation. 

\textbf{State tokens ($S_t$)} provide a dynamic representation of progress at each step $t$. The state tokens are formed from the three orthographic views of the partially constructed model $V_{c,t}$, and the current sketch canvas $V_{sk,t}$ (Fig.~\ref{fig:model_architecture}b). The sketch canvas corresponds to the active sketch in Fig.~\ref{fig:cad_step}. It is an image of the sketch drawn up to the current step, centered on a white background using black strokes, and of the same size as the images $V_{c,t}$. Each image is encoded, and the four embeddings are arranged in a fixed order to form the state token group:
\begin{equation}
S_t = \{V_{c,front,t}, V_{c,top,t}, V_{c,left,t}, V_{sk,t}\}.    
\end{equation}
% Each embedding is added to the same positional embedding $P_{image,i}, i \in [0, 3]$ as the prefixes to distinguish its role (Fig.~\ref{fig:model_architecture}f1). 
The sketch canvas is updated after every sketch operation and reset after any other operation.

A learned positional embedding $P_{image,i}, i \in [0, 3]$ is added to each token in $T_{p}$ and $S_t$ to distinguish its position within the group (Fig.~\ref{fig:model_architecture}f1). 

% This unified approach allows the model to efficiently relate the target geometry to its incremental progress within a consistent representation. 

% Note that we did not use convolution as the visual encoder like the Decision Transformer, as our experiment did not yield improved performance.

\paragraph{Action Tokens ($A_t$)}
The action token $A_t$ is the embedding of the CAD action $a_t$, which consists of a discrete command type $c_t\in[0, K_{cmd}]$ and a parameter vector $\mathbf{p_t}\in[-1,255]^{20}$. There are $K_{cmd}=12$ unique command types, including ``SOL'' and ``EOS'' indicating the start and end of the sequence. To create a unified token, we follow DeepCAD \cite{Wu2021DeepCAD} in normalizing and discretizing all continuous parameters into 256 quantized levels, representing $\mathbf{p_t}$ as a fixed-length vector of 20 integers (using -1 for unused parameters). See our Technical Appendix for more details. The maximum length of the sequence $N_{c}=64$. We create separate embeddings for the command type and the parameter vector (Fig.~\ref{fig:model_architecture}d). The $e_{cmd,t}$ is a $d_m$ embedding of $c_t$. The parameter embedding $e_{params,t}$ is generated by embedding each of the 20 integer values, concatenating the results, and passing it through a linear projection to $d_m$. The two embeddings are then summed and normalized to produce the final action token: 
\begin{equation}
A_t = \mathrm{LN}(e_{cmd,t} + e_{params,t}) \in \mathbf{R}^{d_{m}}.    
\end{equation}

% To increase the information density and reduce the operations that do not alter the state or reward, we remove all intermediate markers ``SOL'' and ``EOS'' from the CAD operation sequences, retaining only the initial ``SOL'' and the final ``EOS''.

\paragraph{Return-to-Go Tokens ($\hat{R_t}$)}
Referring to Decision Transformer, the return-to-go token $\hat{R_t}$ represents the sum of future rewards: $\hat{r_t} = \sum_{k=t}^{L} r_k$, where $L$ is the final step. Given our reward definition, this sum is the total IoU gain to be achieved, representing how far the reconstructed model is from the ground truth. Specifically, $\hat{r_t} = \mathrm{IoU}(M_{L}, M_{target}) - \mathrm{IoU}(M_t, M_{target})$ (Fig.~\ref{fig:model_architecture}c). This scalar value is linearly projected to the model's embedding dimension $d_{m}$ and layer-normalized to form the token:
\begin{equation}
\hat{R_t} = \mathrm{LN}(\mathrm{Linear}(\hat{r_t})) \in \mathbf{R}^{d_{m}}.    
\end{equation}

% During training, we calculate the reward based on the Intersection over Union (IoU) between $M_t$ and $M_{target}$.

% \paragraph{Sequence Assembly}
% The input sequence is constructed by first placing the target prefix tokens $T_{p}$ at the beginning. Following this, the main body of the sequence consists of $N_c$ groups of tokens. These $(\hat{R_t}, S_t, A_t)$ groups are ordered according to their positions in the CAD modeling sequence. We add different group-wise learned positional embeddings $P_{R,t}, P_{S,t}, P_{A,t}$ to the tokens in groups (Fig.~\ref{fig:model_architecture} f2-4). This not only provides group number information, but also distinguishes different types of tokens more clearly.

% The very first group of tokens serves as a start-of-sequence indicator. It is composed of embeddings derived from three orthographic views of an empty rendered environment, a full black sketch image, and a ``SOL'' command token for $A_0$, with return-to-go 1. This complete token sequence, $T_{final} = [T_{p}, (\hat{R_0}+P_{R,0}, S_0+P_{S,0}, A_0+P_{A,0}), \ldots, (\hat{R_{N_{c}-1}}+P_{R,N_{c}-1}, S_{N_{c}-1}+P_{S,N_{c}-1}, A_{N_{c}-1}+P_{A,N_{c}-1})]$ is then fed into the Transformer. This structured input explicitly relates the current and target states, actions, and desired rewards. The model is guided to generate CAD operation sequences that progressively transform the initial empty state towards the geometry depicted in the target views. The reward further incentivizes the model to improve geometric alignment at each step.

\subsection{Training Procedure}
The core of our model is a GPT-style Transformer trained with teacher forcing. The complete sequence is $T_{final} = [T_{p}, (\hat{R_0}+P_{R,0}, S_0+P_{S,0}, A_0+P_{A,0}), \ldots, (\hat{R_{N_{c}-1}}+P_{R,N_{c}-1}, S_{N_{c}-1}+P_{S,N_{c}-1}, A_{N_{c}-1}+P_{A,N_{c}-1})],$ with different learned group-wise positional embeddings $P_{R,t}, P_{S,t}, P_{A,t}$ (Fig.~\ref{fig:model_architecture} f2-4), distinguishing different groups and token types. These groups are ordered according to their positions in the modeling sequence. The very first group of tokens serves as a start indicator. It is composed of embeddings derived from three orthographic views of an empty render environment, a full black sketch image, and a ``SOL'' command token for $A_0$, with return-to-go 1. After the input sequence passes through the Transformer layers, the output representation corresponding to the State tokens is fed into two separate MLP decoders (Fig.~\ref{fig:model_architecture}), which predict the command types and parameters, respectively. We utilize the State outputs for prediction — consistent with the DT framework — under the hypothesis that this encourages the model to more effectively exploit the visual information. We adopt a composite cross-entropy loss similar to DeepCAD \cite{Wu2021DeepCAD}, which is a weighted sum of the losses for both command types and their parameters. 

% : one predicts the probability distribution $\hat{P}(c_{t+1} | S_{t}^{'})$ over the discrete command types, and the other regresses the probability distribution of every single parameter.

\subsection{Inference Procedure}
During inference, the model generates CAD sequences autoregressively. Initially, the orthographic three-view embeddings of the target model and the start indicator group are provided as input. 
As shown in (Fig.~\ref{fig:model_architecture}e), if $a_{t+1}$ is a sketch command, only the sketch component of the state ($V_{sk,t+1}$) is updated, while the 3D model views ($V_{c,t+1}$) and the reward remain unchanged. Conversely, if $a_{t+1}$ is a feature command, the sequence of operations generated so far is executed using a CAD kernel to produce the updated model $M_{t+1}$. This new model is then used to render the new orthographic three-view $V_{c,t+1}$ and $V_{sk,t+1}$ would be set to black image. Since only the three-view of the target ($V_{target}$) is available but not $M_{target}$ during inference, the reward is obtained using the perceptual hashing (pHash) \cite{phash}, which is a common algorithm to analyze the similarity of images. We set the return-to-go during the inference as:
\begin{equation}
\hat{r_{t+1}} =\frac{1-pHash(V_{target}, V_{c,t+1})}{1-pHash(V_{target}, V_{c,0})}.
\end{equation}
$pHash$ represents the hash and Hamming distance calculations; the similarity of the three views is averaged to obtain $pHash$. Since $pHash$ has the same range $[0,1]$ as IoU, this will simulate the training-time return-to-go during the inference, guiding the model to create results completely consistent with the targets. We still leave the reward design during training and inference as an open question, and for now we only use computationally simple and easy-to-implement approaches. The above process repeats, until an ``EOS'' token is generated or a maximum sequence length is reached.

% For example, calculating IoU is simpler than the Chamfer Distance (CD) that requires sampling.

% Concurrently, the return-to-go is computed as the change in IoU: $\hat{R_{t+1}} = 1 - \mathrm{IoU}(M_{t+1}, M_{target})$, as we treat $\mathrm{IoU}(M_{L}, M_{target}) = 1$ during inference time. 

\section{Experiments}

\subsection{Experimental Setup}

\paragraph{Datasets}
Our base model is trained on the CADParser dataset \cite{Zhou2023CADParser}, using 39,430 sequences after removing those failed to render stepwise views. CADParser was selected for its longer sequences and richer command diversity. For broader comparisons, we also used DeepCAD dataset \cite{Wu2021DeepCAD}. All visual inputs are at a resolution of 384x384. The OpenCascade \cite{OpenCascade} was used to render orthographic views and calculate IoU values. The sketch canvases are drawn using Matplotlib.
The arrays of stepwise orthographic views, sketch images, and IoU values are precomputed prior to training.

% The orthographic views are rendered using the OpenCascade \cite{OpenCascade}, while the sketch canvases are drawn using Matplotlib. 
% The IoU values are also computed using the OpenCascade library. 

\paragraph{Implementation Details}
All experiments were conducted on one NVIDIA RTX A6000 GPU. Our model is built upon a Transformer with 8 layers, 12 attention heads per layer, and a feed-forward network dimension of 3072. The embedding dimension $d_m$ for all tokens is 768, resulting in about 65M parameters excluding the visual encoder.
Following the practice of DT, we use the GeLU activation function within the Transformer layers and ReLU for all other nonlinearities. We employed a cosine learning rate scheduler with a linear warm-up of 1000 steps. The learning rate was set to 1e-4, and the base model was trained for 30 epochs on the CADParser dataset \cite{Zhou2023CADParser}. All the visual tokens are given by the same pretrained ViT-Base vision encoder of patch size 32x32, with all parameters frozen. 

% We further discuss hyperparameters in the Technical Appendix. 

\paragraph{Metrics}
We adopt three standard metrics from prior works \cite{Chen2024CADCrafter, Wang2025CADGPT}. The geometric similarity between reconstructed and ground-truth models is measured using Chamfer Distance (CD), which quantifies surface deviation, and Intersection over Union, which assesses volumetric overlap. To evaluate the practicality of the generated sequences, we use the Invalid Rate (IR), which measures the proportion of sequences leading to CAD kernel error. Note that all CD values reported have been multiplied by $100$.

% A lower CD and a higher IoU indicate a more accurate reconstruction. 
% Together, these metrics provide a comprehensive  assessment of the geometric accuracy and the reconstruction efficiency of each method.

\paragraph{Baselines}

We compared our method with different existing methods. CADParser reconstructs CAD sequences through autoregression, conditioned by B-Rep embeddings using cross-attention and Encoder-Decoder Transformers \cite{Zhou2023CADParser}. DeepCAD \cite{Wu2021DeepCAD}, a foundational work in this field, holistically recovers the modeling sequence from a latent vector. We trained adaptive layers to map the three-view embeddings to its latent space, enabling conditional generation. SkexGen \cite{Xu2022SkexGen} generates command sequences through autoregression. We modified and retrained its codebook selector to enable selection based on the three-view embeddings. HNC-CAD generates the entire B-Rep model holistically without requiring a command sequence \cite{Xu2023HNCCAD}. We trained adaptive layers to map the three-view embeddings to the latent space of its model encoder. More about baseline selection could be found in the Technical Appendix.
All three-view embeddings are obtained using a pretrained ViT encoder similar to our setup.

% Its pretrained code-tree generator and model generator are retained.

% \begin{itemize}
%     \item CADParser reconstructs CAD sequences through autoregression, conditioned by B-Rep embeddings using cross-attention and Encoder-Decoder Transformers \cite{Zhou2023CADParser}. 
%     \item DeepCAD \cite{Wu2021DeepCAD}, a foundational work in this field, can holistically recover the modeling sequence from a learnable latent space vector. We trained adaptive layers to map the three-view embeddings to the DeepCAD latent space, enabling conditional generation. 
%     \item SkexGen \cite{Xu2022SkexGen} generates command sequences through autoregression. We modified and retrained its codebook selector to enable selection based on the three-view embeddings.
%     \item  HNC-CAD is a representative work that can generate the entire B-Rep model holistically without requiring a command sequence \cite{Xu2023HNCCAD}. We trained learnable adaptive layers to map the three-view embeddings to the latent space vector of its model encoder. Its pretrained code-tree generator and model generator are retained.
% \end{itemize}

\begin{figure*}[t]
    \centering
    \includegraphics[width=0.75\textwidth]{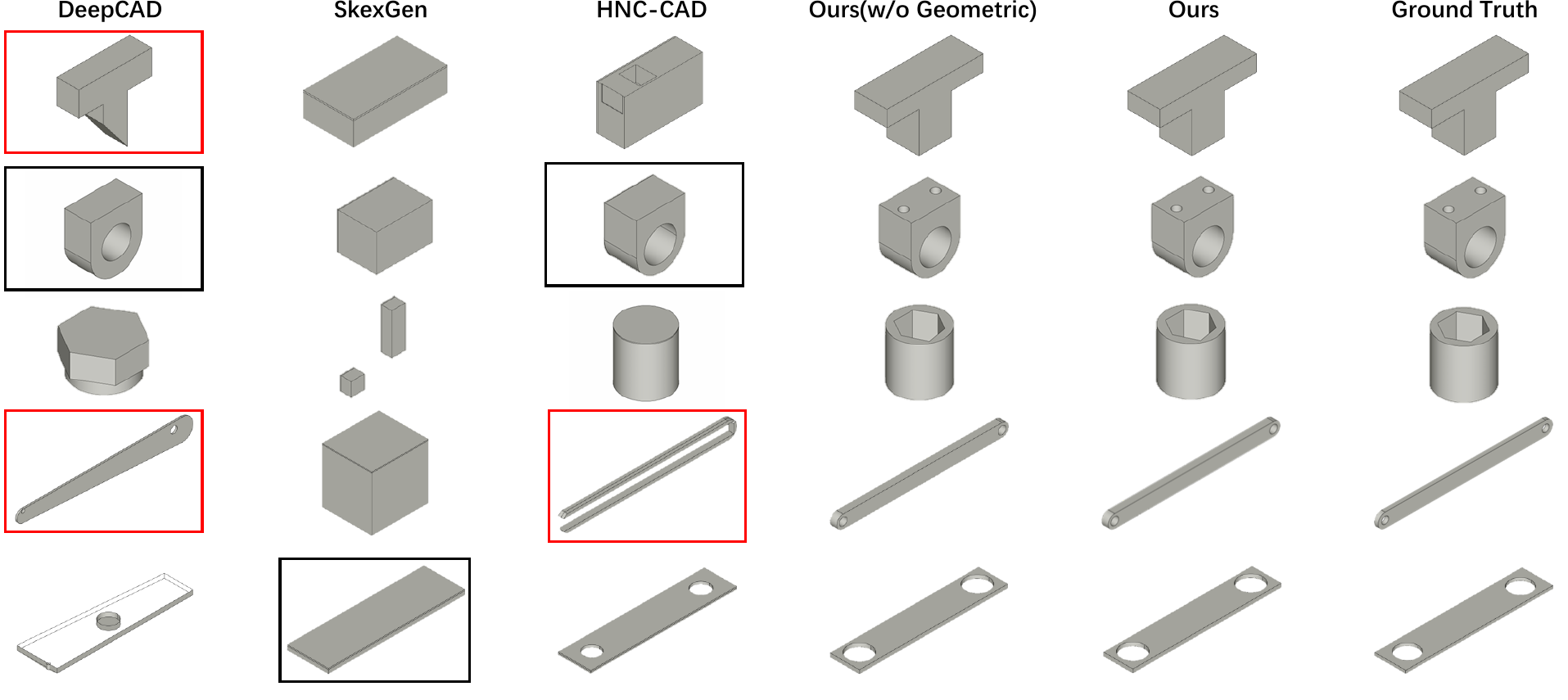}
    % \fbox{\rule[-.5cm]{0cm}{8cm} \rule[-.5cm]{16cm}{0cm}}
    \caption{Results of reconstructing CAD modeling sequences using orthogonal three-view images on the DeepCAD test set, comparing ours with other methods. Some typical errors are highlighted. Red boxes indicate incorrect features, such as incorrect holes or unnecessary cuts. Black boxes indicate missing features, where necessary details are missing.}
    \label{fig:qualitative_double}
\end{figure*}

% \begin{figure}[h]
%     \centering
%     \includegraphics[width=\columnwidth]{images/fig3-cropped.pdf}
%     % \fbox{\rule[-.5cm]{0cm}{6cm} \rule[-.5cm]{8cm}{0cm}}
%     \caption{Reconstruction results on real-life photos.}
%     \label{fig:qualitative_single}
% \end{figure}

\subsection{Quantitative Results}

\paragraph{Results on CADParser}
We first evaluate our base model, trained exclusively on the CADParser dataset. We compare our performance against the original value reported in the CADParser paper, since their code is not publicly available. We evaluate the performance with and without geometric input. Without geometric means $V_{target}$ is the only condition and the reward is calculated from pHash; conversely, having geometric means the IoU can be calculated during inference. As shown in Table~\ref{tab:cadparser_comparison}, our method outperforms CADParser, and the non-geometric version also achieves strong performance, highlighting the robustness of our mechanism. With B-Rep embedding inputs, CADParser still cannot do better in geometric consistency, and the B-Rep embedding limits its extendability. Interestingly, the non-geometric version achieves a lower IR, likely because pHash is more robust than IoU, suggesting potential for future reward design.

\begin{table}[h]
\caption{Comparison with CADParser on its dataset.}
\centering
\begin{tabular}{lccc}
\toprule
Method      & Median CD $\downarrow$ & IoU $\uparrow$ & IR $\downarrow$ \\ \midrule
CADParser & NA  & 0.81  & NA   \\
Ours (w/o geometric) & 0.29 & 0.84 & \textbf{7.8\%} \\
Ours (w/ geometric) & \textbf{0.23}  & \textbf{0.85}  & 9.7\%   \\ \bottomrule
\end{tabular}
\label{tab:cadparser_comparison}
\end{table}

\paragraph{Results on DeepCAD}
To compare against additional baselines trained on DeepCAD, we randomly sampled 50\% of the dataset, yielding 84,197 sequences. We split the sampled data evenly into two halves: the first was merged with CADParser data for pretraining, and the second was used for fine-tuning. This strategy balanced data usage with training efficiency. The model Ours* in Table~\ref{tab:deepcad_comparison} shows the result. Additional results for different mixing ratios are provided in our Technical Appendix. Our method achieves the best performance across all metrics. In terms of median CD, it reduces the error by nearly 15\% compared to the second-best, despite using only half of the DeepCAD dataset, whereas other methods are pretrained on the full dataset. This highlights the geometric accuracy, robustness, and data efficiency of our approach. Although SkexGen adopts a complex codebook and separates sketching and extrusion, it struggles to preserve high-density geometric information across three views. HNC-CAD yields suboptimal geometric consistency and its highest failure rate suggests that holistic B-Rep reconstruction can compromise robustness despite increased model complexity. 

% Compared to DeepCAD, our method shows a clear advantage, while SkexGen even underperforms DeepCAD in geometric accuracy. 
% Overall, our results confirm the effectiveness of stepwise visual feedback for achieving both accuracy and resilience.

\begin{table}[h]
\caption{Comparison with baselines on the DeepCAD dataset.}
\centering
\begin{tabular}{lccc}
\toprule
Method    & Median CD $\downarrow$ & IoU $\uparrow$ & IR $\downarrow$ \\ \midrule
DeepCAD   & 1.18  & 0.76  & 13.6\%   \\
SkexGen    & 2.93 & 0.68  & 11.5\%     \\
HNC-CAD    & 0.44  & 0.78  & 18.9\%   \\
Ours*(w/o geometric) & \textbf{0.37} & 0.83 & 7.4\% \\
Ours*(w/ geometric)    & 0.38  & \textbf{0.84}  & \textbf{7.3\%}   \\ \bottomrule
\end{tabular}
\label{tab:deepcad_comparison}
\end{table}

\subsection{Ablation Study}

To understand the contribution of each component in our framework, we conducted a series of ablation studies on the CADParser dataset. The results are summarized in Table~\ref{tab:ablation_study}.

\begin{table}[h]
\caption{Ablation study of our model's components.}
\centering
\begin{tabular}{lccc}
\toprule
Method                      & Median CD $\downarrow$ & IoU $\uparrow$ & IR $\downarrow$ \\ \midrule
w/o Stepwise         & 0.31  & 0.83  &  \textbf{3.5\%}  \\
w/ Single-View              & 0.50  & 0.81  & 17.6\%   \\
w/o Sketches                & 1.42  & 0.74  &   33.4\%  \\
w/o Reward                  & 0.25  & 0.84  & 5.1\%   \\ 
\midrule
Ours           & \textbf{0.23}  & \textbf{0.85}  & 9.7\%   \\ 
\bottomrule
\end{tabular}
\label{tab:ablation_study}
\end{table}

Removing the stepwise visual and reward feedback entirely (``w/o Stepwise''), which reduces the model to training only on the prefix tokens $T_{p}$ and action tokens $A_t$, leads to an obvious drop in performance, emphasizing the importance of continuous visual guidance. Using only a single fixed-angle view for stepwise views (``w/ Single-View'') performs even worse due to the serious self-occlusion, confirming the value of comprehensive visual information. Excluding sketch information (``w/o Sketches'') results in a severe performance decrease. This model was overfitted very soon, indicating that sketches provide valuable guidance, and long-term unchanging states across sketching actions seriously damage model performance. Finally, removing the reward signal (``w/o Reward'') turns a non-RL, strong supervised seq2seq. This causes a slight performance degradation, validating the effectiveness of the RL formulation toward improved geometric alignment.

\subsection{Qualitative Results}

Fig.~\ref{fig:qualitative_double} shows the result on synthetic CAD images rendered using OpenCascade. Our approach faithfully reconstructs fine geometric details, outperforming others in both the with and without geometric settings. Competing methods often fail to preserve structural fidelity. See the Technical Appendix for more comparisons on CADParser and DeepCAD test sets.

% We also observed self-correction phenomena, also documented in the Technical Appendix.

% —DeepCAD produces unclosed geometries, while SkexGen collapses to oversimplified primitives
% Figure~\ref{fig:qualitative_single} shows reconstructions from mobile phone photographs. Without any fine-tuning on real-world imagery, our model captures key object features and produces CAD outputs closely matching the target geometry. Input photos are preprocessed by background removal and color correction. These results underscore our method’s robustness and strong generalization to in-the-wild inputs.

\section{Conclusion}

% While our approach achieves promising results for CAD sequence reconstruction, it has several limitations. First, the reliance on three-view orthographic projections constrains data collection, and it is promising to weaken this constraint or extend to arbitrary viewpoints. Second, the current training-time reward computation depends on the ground-truth CAD model, which restricts inference performance in settings with only visual inputs. Finally, extending the framework from reconstruction to fully unconstrained CAD generation remains an open question.

We have presented SOV-CAD, a novel framework that formulates CAD modeling sequence reconstruction as an offline RL problem. Based on the Decision Transformer, our method leverages rich, stepwise visual feedback from orthographic views to mimic the iterative workflow of human designers. Extensive experiments show that SOV-CAD achieves state-of-the-art performance while maintaining strong generalization and robustness across datasets. 

% Future work could focus on extending to arbitrary viewpoints, optimizing the reward design, and fully unconstrained CAD generation.

% By effectively bridging unstructured visual inputs with structured parametric modeling sequences, our work highlights the promise of sequential decision-making frameworks in tackling complex engineering and design challenges.

\section{Acknowledgment}

This work was supported by "Pioneer" and "Leading Goose" R\&D Program of Zhejiang (2023C01045) and  Zhejiang Key Laboratory Project (2024E10001). We also thank Yijing Chen and Yuntai Bao for their assistance.

\bibliographystyle{IEEEbib}
\bibliography{icme2025references}

\end{document}